\documentclass{article}

\usepackage{arxiv}

\usepackage[utf8]{inputenc} 
\usepackage[T1]{fontenc}    
\usepackage{hyperref}       
\usepackage{url}            
\usepackage{booktabs}       
\usepackage{amsfonts}       
\usepackage{nicefrac}       
\usepackage{microtype}      
\usepackage{lipsum}
\usepackage{graphicx}
\usepackage{soul}
\usepackage{bbold}
\soulregister\cite7
\soulregister\ref7
\soulregister\pageref7
\usepackage{algorithm,algorithmic}
\usepackage{amsmath}

\title{YOLOpeds: Efficient Real-Time Single-Shot Pedestrian Detection for Smart Camera Applications}

\author{
  Christos Kyrkou\thanks{ckyrkou@gmail.com, www.christoskyrkou.com
  } \\
  KIOS Research and Innovation Center of Excellence\\
  University of Cyprus\\
  1 Panepistimiou Avenue, Nicosia Cyprus \\
  \texttt{kyrkou.christos@ucy.ac.cy} \\
}

\begin{document}
\maketitle

\begin{abstract}
Deep Learning-based object detectors can enhance the capabilities of smart camera systems in a wide spectrum of machine vision applications including video surveillance, autonomous driving, robots and drones, smart factory, and health monitoring. Pedestrian detection plays a key role in all these applications and deep learning can be used to construct accurate state-of-the-art detectors. However, such complex paradigms do not scale easily and are not traditionally implemented in resource-constrained smart cameras for on-device processing which offers significant advantages in situations when real-time monitoring and robustness are vital. Efficient neural networks can not only enable mobile applications and on-device experiences but can also be a key enabler of privacy and security allowing a user to gain the benefits of neural networks without needing to send their data to the server to be evaluated. This work addresses the challenge of achieving a good trade-off between accuracy and speed for efficient deployment of deep-learning-based pedestrian detection in smart camera applications. A computationally efficient architecture is introduced based on separable convolutions and proposes integrating dense connections across layers and multi-scale feature fusion to improve representational capacity while decreasing the number of parameters and operations. In particular, the contributions of this work are the following: 1) An efficient backbone combining multi-scale feature operations, 2) a more elaborate loss function for improved localization, 3) an anchor-less approach for detection, The proposed approach called \textit{YOLOpeds} is evaluated using the PETS2009 surveillance dataset on 320x320 images. Furthermore, a real-system implementation is presented using the Jetson TX2 embedded platform. Overall, \textit{YOLOpeds} provides real-time sustained operation of over $30$ frames per second with detection rates in the range of $86\%$ outperforming existing deep learning models.
\end{abstract}


\section{Introduction}
Visual object detection is of one the most fundamental and challenging problems in computer vision. It has received great attention in recent years \cite{DeepObjectDetectionReview:2019,DLCV:IET} as it sets the foundation for enabling a plethora of applications, such as surveillance cameras\cite{SmartCameraas:2011}, unmanned aerial vehicles (UAV)\cite{kyrkou18dronet}, and self-driving vehicles \cite{6033589}. In particular pedestrian detection is an essential and significant task in any intelligent video surveillance system and smart camera environment. In previous years, the works targeting pedestrian detection were mostly based on the sliding window search and the handcrafted features such as Histograms of Oriented Gradients \cite{cascadeHOG:Zhu:2006}. More recent approaches based on deep learning have demonstrated significant progress in terms of accuracy \cite{AlexNet}. Such methods rely on a convolutional neural network (ConvNet) to learn meaningful features for image representation while simultaneously learning to localize the objects within the image by regressing their bounding boxes. Approaches utilizing ConvNets are primarily categorized into two groups, region-based and single shot. The former combine a region proposal method with ConvNet features for detection (e.g., \cite{FasterRCNN:2017}) and outperformed conventional detectors in terms of accuracy. The latter family of methods also referred to as one-stage detectors, such as YOLO\cite{YOLOV2,YOLOV3} and SSD \cite{liu2016ssd}, can further reduce the computations by removing the region proposal stage and formulating object detection as a dense regression problem. However, such approaches are resource- and power- hungry and have not yet matched the efficiency of previous methods \cite{EnergyGap:CNNVHOG:2017}. Consequently, it is challenging to deploy deep visual object detection algorithms on embedded devices due to the trade-off among energy consumption, accuracy, and speed; thus preventing their use in scenarios that impose low-latency constraints during inference.

There are additional problems using existing classification backbones for object detection tasks: (i) involvement of extra stages compared with the backbone network for ImageNet classification, (ii) traditional backbone utilize higher downsampling factors, which is more suited to the visual classification task. However, the spatial resolution is compromised which leads to failures in accurately localizing the large objects and recognizing the small objects. Multi-scale feature fusion is critical to achieving better accuracy for object detection problems. However, multi-branch predictions introduce more complexity as they increase the number of operations and parameters. Hence, they should be used selectively. Overall, smaller ConvNets have many desirable properties such as training faster, small memory footprint, more efficient power consumption, and reduced computational requirements. All these contribute to the realization of always-on vision systems for smart camera applications.

To this end, a ConvNet architecture is proposed suitable for real-time pedestrian detection on embedded smart camera applications. Rather than utilize a pretrained feature extractor, or just change the number of layers and filter sizes in YOLO and SSD architectures, the network backbone is directly designed for inference on embedded smart camera systems by incorporating separable convolutions and successive downsampling. It relies on selective use of dense connections for maintaining information and merges features from different resolutions to regress detection and localization. To simplify the overall framework for pedestrian detection, the anchors are also removed and the ConvNet directly regresses bounding boxes as factors with respect to the image dimensions. Various aspects are analyzed such as detection accuracy, operations, model size, speedup rate, parameter utilization, and memory of various network structures. The proposed convolutional neural network and overall framework maintain the accuracy of high-end implementations while outperforming existing works in terms of inference speed. Specifically, when implementing the proposed approach on an NVIDIA Jetson TX2 embedded platform it can achieve speed of over 30 FPS with an accuracy of 86\% on a pedestrian surveillance image sequence based on the PETS2009 dataset \cite{PETS2009}. The contributions of this work bring \textit{YOLOpeds} to be comparable to high performing object detectors with a significant improvement in inference speed, and efficient utilization of parameters.

The rest of this paper is structured as follows. Section \ref{rw:background} provides the background on convolutional neural network architectures for classification, object detection, and edge processing. Section \ref{methodlogy} provides details on the methodology, ConvNet architecture and techniques used to developing \textit{YOLOpeds}, training process and optimizing loss function for pedestrian detection. Section \ref{experiments} presents an analysis of the different models and techniques as well as evaluation on the Jetson TX2 embedded platform. Finally, Sections \ref{sec:conc} provides concluding remarks.

\section{Background}\label{rw:background}
\subsection{Visual Object Detection with Deep Learning}\label{rw:od}
In the last decade, a lot of progress has been made on ConvNet-based object detection. There are basically two main families of deep object detection models: 1) region-based detectors, consisting of region proposal stage followed by classifier \cite{FasterRCNN:2017}, and 2) single-shot detectors \cite{YOLOV2,YOLOV3,liu2016ssd}, consisting of a single ConvNet to perform object detection. Single-shot detectors such as YOLO (You Only Look Once)\cite{YOLOV2,YOLOV3}, have generally shown significant potential for resource constraint smart camera applications compared to other approaches. 

\subsubsection{Region Proposal Methods}
Detectors belonging to the R-CNN (region-based convolutional neural networks) family fall under two-stage detectors. The process has two stages, first extracting region proposals and then classifying each proposal and predict the bounding box. Two-stage approaches generally lead to good detection accuracy but the inference time of such detectors is heavily constraint by the number of proposals \cite{DeepObjectDetectionReview:2019}. Faster R-CNN is the latest approach in region proposal methods and made further improvements on the region proposal framework such as introducing a region proposal Network (RPN) that shares full-image convolutional features with the detection network, thus enabling nearly cost-free region proposals. An RPN is a fully convolutional network that simultaneously predicts object bounding boxes and scores at each position. RPNs are trained end-to-end to generate accurate region proposals. Because of the region proposal network, it the detector can be trained end to end, and the network is faster than previous iterations. One of the drawbacks of methods such as Faster R-CNN is that they are time consuming in inference phase since they generate many region proposals on the original input image size and need to classify each of the regions which introduces significant delays. Even when restricting the number of regions that need to be processed \cite{Kouris2019InformedRS} the performance improvements are still not enough to facilitate real-time operation in embedded applications. Besides, since each region is classified individually there is lack of context within the network. It is why researchers have also been interested in single-stage methods, especially in the embedded domain. The focus of this paper concerns single-shot networks due to their significantly higher accuracy-performance trade-offs which makes them more suitable for smart camera applications. Hence, comparisons and analysis will be restricted within the family of single-shot detection algorithmic domain.

\subsubsection{Single-Stage Detectors}\label{rw:ssd}
Single-shot methods \cite{YOLOV2,YOLOV3,liu2016ssd} cast object detection from a pure classification problem to a combined regression and classification problem. Instead of having two networks for two different tasks (region proposal, and region classification), a single ConvNet is employed for both tasks simultaneously. A single neural network predicts bounding boxes and class probabilities directly from full images in one evaluation. Since the whole detection pipeline is a single network, it can be optimized end-to-end directly on detection performance. The unified architecture is extremely fast but generally is not as accurate as Faster R-CNN. However, all object detection predictions for a single image are made in a single forward pass, compared to hundreds to thousands of passes that need to be performed to get the final results for region proposal-based networks. Hence, detectors belonging to the single-shot family are more likely to run faster than the two-stage detectors since all computations are in a single network. This alleviates the need to separately optimize the performance of the generator and then the detector which introduces many iteration cycles during training. As such, YOLOv3 \cite{YOLOV3} and SSD\cite{liu2016ssd} are currently considered the state-of-the-art for single-shot object detection.

Single-shot detectors basically receive an input image, resize the image into a predetermined size, and generate bounding boxes and class probabilities for a certain number of objects along a grid that is spatially associated with the input image. A main similarity across both region- and single-shot- based approaches is the use of anchor boxes \cite{YOLOV2,FasterRCNN:2017} which are predetermined set of boxes with particular height/width ratio based on the ground truth bounding boxes of a dataset, that act as template bounding boxes. Anchors cover spatial position, scales, and aspect ratios across an image without the need for an extra branch for extracting region proposals. Using the ground truth bounding boxes of the dataset and K-means clustering, anchor boxes can be extracted which serve as a prior for the predicted bounding boxes. These boxes can be used to calculate the average recognition values for a specific object. In this instance, the regressor effectively learns to predict a multiplicative factor on these aspect ratios and prior sizes. This aids the learning processes especially for datasets with many objects and high varying aspect rations \cite{YOLOV2}. The number of boxes regressed depends on the number of grids in each detection head, and the dimensions of the grid. Often multiple detection heads are extracted from different network layers to better detect objects are multiple resolutions. Each detection head may have a different grid size and subsequently regresses a different number of bounding boxes. For example SSD regresses bounding boxes at different scales (e.g., $1\times1,4\times4, 8\times8, etc.$), while a similar approach is employed in YOLOv3\cite{YOLOV3} but not YOLOv2 \cite{YOLOV2}. Regression across different layers can increase the potential for noisy detections, incur higher cost on post processing and generally increase the computational demands and network working memory. The main differences between SSD and YOLO methods relate to the backbone networks, loss functions and regression formulation. More efficient versions of SSD utilize the MobileNetV2 architecture \cite{MobileNetV2:2018} while YOLO methods use a convolutional network called DarkNet \cite{YOLOV2} with different number of convolutional layers ranging from $19$ to over $50$. As such, all object detection predictions for a single image are made in a single forward pass, compared to hundreds to thousands of passes that need to be performed to get the final results for region proposal-based networks. This makes the single-shot family of network architectures significantly faster to run, and thus better suited as a starting point for embedded object detection.

\subsection{Detection Meta-Architectures}\label{rw:arch}
The detection frameworks outlined in the previous section use a common practise of directly fine-tuning from ImageNet pre-trained models designed for image classification in a process called transfer learning. VGG \cite{VGG.2014} first investigated deeper network structure and showed that using successive $3\times3$ convolutions can produce better results than single layers with larger filters. ResNet \cite{ResNet.2015} is a $152$ layer deep network that tried to solve the problem of vanishing/exploding gradients for deeper networks by learning residual representation functions enabling the training of deeper networks. It is widely used as a feature extractor along with its smaller variants such as \textit{ResNet50}. DenseNets take the idea of residual connections and expand on it by proposing a simple strategy of concatenating the output of each layer with the output from all previous layers and then propagating it to the next layer. This approach also improves the overall gradient flow and can lead to decrease in number of parameters and more diversified features. The MobileNet \cite{MobileNetV2:2018} family of networks use depthwise separable convolution, batch normalization and ReLU in an efficient ConvNet architecture for embedded vision applications. The main benefit from this architecture is the reduction in number of parameters. These architectures are geared towards generic vision tasks often overlooking their impact and requirements for resource-constraint applications.  However, deep network architectural issues must be considered to achieve the optimal trade-off between accuracy and speed with minimal computational and memory cost.

\subsection{Pedestrian Detection}
Pedestrian detection on its own has received significant interest from the computer vision community. The work in \cite{multicue:cvpr:2009} explores the combination of multiple and complementary feature types to help improve performance in a sliding-window framework. Unsurprisingly experiments indicate that incorporating motion information improves detection performance significantly. Such techniques can be considered complementary to more traditional pattern recognition based techniques. The majority of approaches such as in \cite{OccPedDet:CVPR2018}, utilize the Faster R-CNN framework due to its higher accuracy but do not considering embedded applications and constraints. In \cite{Zhang:ECCV:2016} the authors consider the use of Faster R-CNN framework for pedestrian detection and address its problems with respect to instance detection and hard-negative mining. They apply Region Proposal Networks (RPN) with a boosted forests classifier to address these detection issues. In \cite{FusedDNN:2017} the fusion of multiple network outputs is proposed where an SSD network first generates all possible pedestrian candidates of different sizes and occlusions followed by multiple deep neural networks that operate in parallel for further refinement of these pedestrian candidates. The inclusion of multiple networks makes it increasingly difficult for deployment in smart camera settings especially since evaluation is performed on an NVIDIA Titan X GPU. However, the main limitation of region proposal methods which is the iterative classification of many proposals is not addressed. The authors in \cite{YPD.2018} introduce the variant of YOLOv2 referred to as \textit{Y-PD} with some modifications geared towards pedestrian detection. It makes use of larger filters and underlying loss function is identical to YOLOv2. Essentially, it is a direct re-purposing of YOLOv2 for the pedestrian detection task. The particular network is deployed on desktop GPU computer without considering smart camera applications.

\subsection{ConvNets at the Edge}
In typical application scenarios, convolutional neural networks run on powerful GPUs which have sufficient computing and memory resources. In light of the excessive resource demands of modern ConvNets, a common approach is to use powerful cloud datacenter for training and evaluating them. Input data generated from edge devices, such as smart cameras, are sent to the cloud for processing, and the results are sent back to those devices after the inference. There is also the option of a hybrid solution where models can be broken up so that most of the processing is done on-device, with the final stages of computation being sent to the cloud for processing. However, sending data to the cloud for processing also requires an active and reliable internet connection and bandwidth. This is a significant barrier to access for remote and developing areas. In addition, there is always the privacy and security aspect when sending data which can be compromised. Avoiding heavy data processing between devices and the cloud can also be a huge cost-saver. With the advancement of the technology and the powerful devices such as Jetson by Nvidia\cite{NVIDIA:Jetson:TX2} that can analyze real-time data at the edge, a new wealth of possibilities opens up for potential applications for architectures and algorithms that can take maximum advantage of such hardware. However, it is often prohibitive to use typical neural network architectures in such settings due to the increased computational and memory requirements. Especially, since the model size is huge and cannot facilitate on-chip local storage which is critical for low-cost and low-power applications. Other approaches that consider the pruning of deep learning networks \cite{Keeffe:PrunedOD:2018} for improved efficiency can result in fewer trainable parameters and lower computation requirements in comparison to the original models, however, the accuracy is reduced and retraining is often necessary. There are two main drawbacks to existing approaches, first, optimizing already large networks through pruning may not lead to the desired accuracy-speed trade-offs as the margin for improvement is small, and second, transferring weights from generic object may not translate to good detection performance on more task-specific settings. To cope with such issues it is convenient to explore task-specific network architectures \cite{ScratchDet:2018}. 

\begin{figure*}[t]
\centering
\includegraphics[width=1\linewidth]{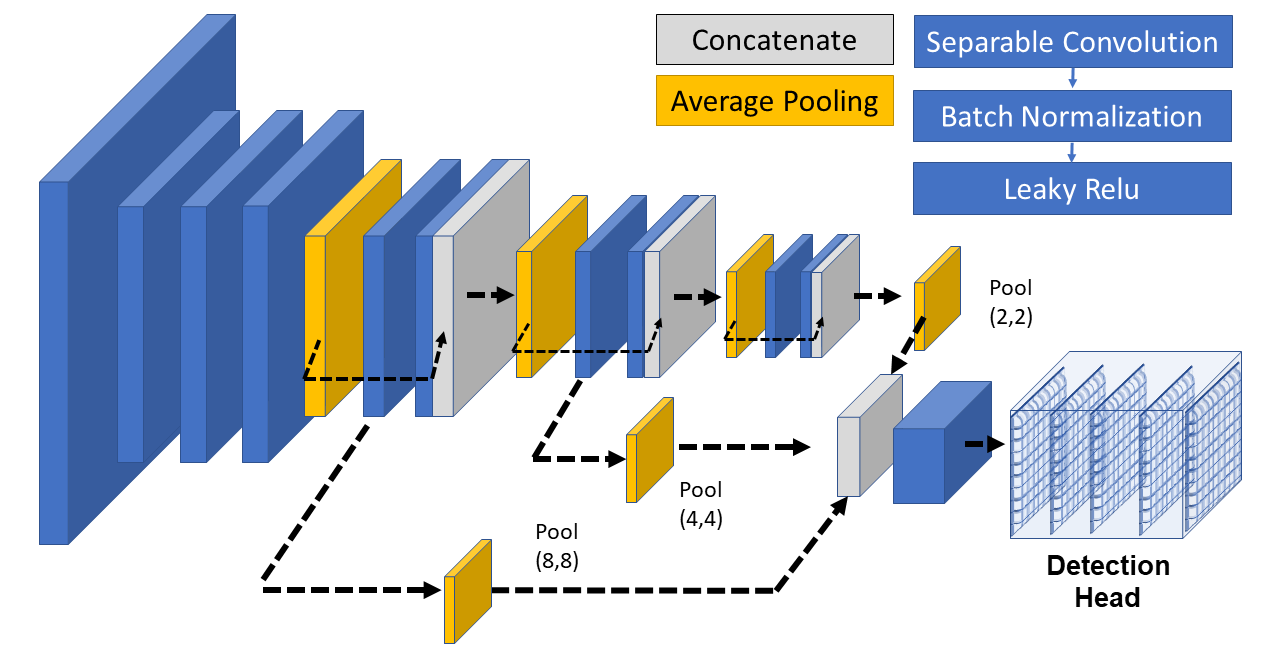}
\caption{\textit{YOLOpeds} feature extractor: A Deep ConvNet with short and long connections and multi-scale information fusion for detection.}
\label{fig:YOLOpeds}
\end{figure*}

\section{Methodology}\label{methodlogy}
Lightweight object detection is still a challenge and there is a need to optimize deep learning networks to speed up the detection algorithm so that it can run smoothly on mobile devices. For this reason, purpose built ConvNets such as the face detection system in \cite{CNNFace:Haoxiang:2015}, and vehicle detection system \cite{kyrkou18dronet} demonstrate that they can provide better accuracy/performance trade-offs for edge processing through the design of powerful but light-weight networks. To this end, this work goes beyond just modifying an existing approach, by changing the philosophy of the current network towards embedded applications of pedestrian detection. Inspired by the good practises from previous works outlined in Section \ref{rw:arch}, \textit{YOLOpeds} is introduced in an attempt to address the discrepancy from classification backbones to detection networks. At a glance the proposed architecture features the hierarchical processing with small-sized filters, separable convolutions to reduce parameter count and more efficient computation, short connections between layers and concatenation to build a more diverse set of features, and longer connections to perform detection using multi-scale information. The above design principles are combined so as to increase the effectiveness both in terms of accuracy and representation, and inference speed in order to get real time pedestrian detection in low-cost low-power environments. In addition, the loss function is modified to include an Intersection-over-Union (IoU) component to improve localization. To also simplify the overall detection framework an approach is proposed that does not use anchor-boxes for steering the bounding box prediction. Each major feature is elaborated next.

\subsection{Feature Extraction}
At feature extraction stage, a neural network extracts the image features in a hierarchical manner, the quality of which is critical to the detection accuracy. The network architecture also has an impact on inference performance in terms of throughput. In addition, smaller networks exhibit better memory management and consequently improved power consumption \cite{DATELowPowerImage2018}. Given the significant resources consumed by the feature extraction backbone network, it is crucial to design a powerful but light-weight network. Backbone features that have worked well in previous works are investigated and adapted accordingly to the pedestrian detection problem. The \textit{YOLOpeds} backbone is comprised of $10$ convolutional layers. All convolutional layers are implemented using separable convolutions that factors standard convolution into a depth-wise and point-wise convolution. Depthwise convolutions apply a single filter per input channel (input depth) while pointwise convolutions create a linear combination of the output of the depthwise layer using $1\times1$ convolution. This results in reducing the parameters by a factor analogous to the depth of the output feature map and the size of the kernel, at only a small reduction in accuracy \cite{MobileNetV2:2018}. 

A key intuition behind the arrangement of the convolutional layers is the fact that aggressive downsampling can lead to loss of localization information which is critical for detection networks. As such, within the first block (first four conv layers in Fig. \ref{fig:YOLOpeds}) the initial convolutional layer only downsamples the input image by a factor of $2$ using strided convolutions. Then the subsequent $3$ convolutional layers operate at the same scale. 

The first convolutional layer has $16$ kernels, while the latter two have $32$. All convolutional layers in this block operate with a $5\times5$ filter size. Keeping the number of filters low critical for achieving higher performance at this point where the feature map resolution is not low enough. The rest of the convolutional layers employ $3\times3$ kernels with increasing number of filters of up to $256$. 

The backbone employs dense-like connections where previous layers are concatenated with deeper layers to permit for gradients to flow better during back propagation and builds more diverse set of features. Layer concatenation happens at $3$ stages in the backbone architecture, every two convolutional blocks. The merging of feature maps happens by concatenating the pooling layers with the second following convolutional layer. This serves the purpose of compensating for the pooling which reduces the spatial resolution by propagating and fusing additional information with less semantic meaning but lower-level features. Average pooling was preferred instead of Max pooling commonly used in classification networks, as it empirically performed better. This can be attributed to the fact that it better preserves spatial information which is critical for good localization in object detection networks. Finally, it is important to note that \textit{YOLOpeds} is inspired by DenseNets but does not follow the same dense connection pattern of connecting a layer with all its previous layers as this progressively increases the computation and memory demands. In contrast a selective use of short and long connections is used to provide a trade-off between the added computation and the use of more diverse feature information.

The feed-forward backbone network downsamples the input image by a factor of $32$. Such coarse feature maps are incapable to accurately provide competitive accuracy in an efficient manner; especially for the dense and occluded objects as in Fig. \ref{fig:detres}. Thus, as shown in Fig. \ref{fig:YOLOpeds}, the network has multiple layer outputs that are combined together prior to the detection block that actually regresses the bounding boxes. Instead of upsampling the high-level feature maps and predicting multiple locations from different branches as in Feature Pyramid Networks \cite{FPN:2016}, they are combined with the low-level ones through channel concatenation after further downsampling through average pooling. Unlike element-wise addition, channel concatenation provides a learnable way to combine feature maps at different levels by enriching semantic information and detailed location information to localize the object of interest. As such, the overall architecture is more akin to YOLOv2 rather than YOLOv3 since there are no multiple output branches of different scales and grids which leads to a simpler training process. Instead the architecture learns multiple object sizes through specialization during training, more details are provided in Section \ref{detectionHead}. Overall, the \textit{YOLOpeds} backbone provides lighter computation load, as well as a rich set of feature hierarchies. Other methods rely on large batch sizes and the effects of batch normalization to provide from scratch training \cite{ScratchDet:2018}.

\subsection{Detection Head}\label{detectionHead}
The detection head receives the output of the feature extraction backbone in order to regress the object probabilities and bounding box coordinates. Since the pedestrian detection task for smart camera and surveillance applications is more streamlined and objects do no exhibit such high variability in terms of sizes and aspect ratios a mechanism is proposed for regressing the bounding boxes without reliance on anchor boxes to simplify the training process. The additional advantage is that the model is then not sensitive to anchor boxes and the different algorithms used to find them.

\begin{figure}[t]
\centering
\includegraphics[width=0.7\linewidth]{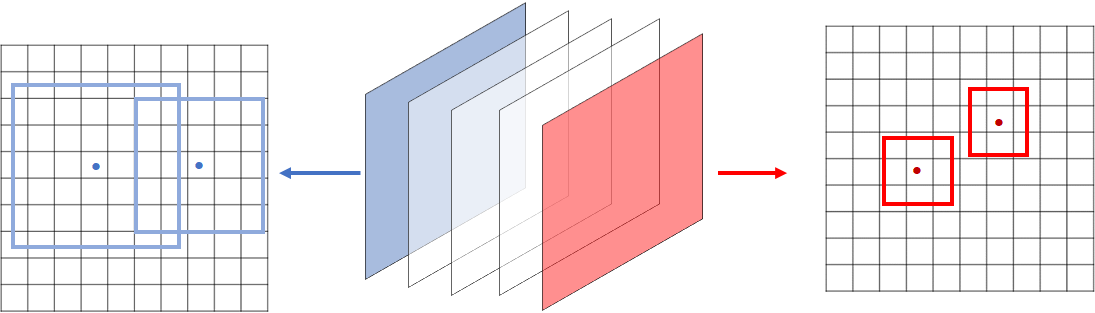}
\caption{Each grid in the detection head is responsible for predicting boxes of different scales, corresponding to objects with increased IoU over the image size.}
\label{fig:ScaleHeads}
\end{figure}

\begin{algorithm}[t]
\caption{Pseudo-code for Bounding Box Responsibility Selector}
\label{alg:bboxselect}
\textbf{Inputs: }$GRIDS,bbox,image$\\
\textbf{Output: }$responsibility_selector$
\begin{algorithmic}[1]
\STATE \textbf{function} \textit{assignGrid}(bbox,image,GRIDS)\:
\FOR{bbox in image}
\STATE $boxAREA = \sqrt{IoU(bbox,image_{size})}$
\FOR{grid in GRIDS}
\STATE $gridAREA\leftarrow getGridResponsibleArea(grid)$
\IF{boxAREA < gridAREA \textbf{and} grid \textit{is} \textit{not} \textit{occupied}}
\STATE selector(grid,bbox) = 1
\STATE \textit{break}
\ENDIF
\ENDFOR
\ENDFOR
\STATE \textbf{return} \textit{selector}
\end{algorithmic}
\end{algorithm}

\begin{figure}[t]
\centering
\includegraphics[width=0.7\linewidth]{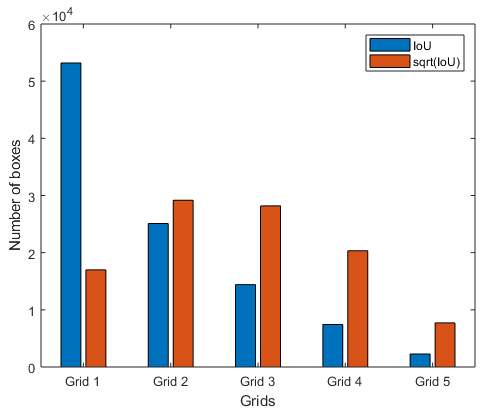}
\caption{Grid assignment for each bounding box based on the area it occupies within the image, reflected by the IoU with the input image. Selecting the grid that each box will be assigned to based on the square root of its IoU in the input image results in a more uniform distribution. (Best viewed in color)}
\label{fig:boxes_assign}
\end{figure}

Inspired by the You Only Look Once (YOLO) algorithm, \textit{YOLOpeds} directly learns the object presence probability and the bounding box coordinates without anchor reference. This is achieved by predicting size-specific boxes as shown in Fig. \ref{fig:ScaleHeads}. in particular each grid is responsible for learning size-specific detections and bounding boxes following the procedure outlined in Algorithm \ref{alg:bboxselect}. A key point here is the assignment of ground truth boxes to predictors in each grid. YOLO first identifies the center of each box and then selects the anchor that best fits the bounding box. Since the aim is to provide anchor-less prediction the assignment is done in a different way using the size of the object as a guide. The area that the object occupies within the image determines the grid that will be responsible for predicting its presence and size. In this way smaller objects are assigned to first grid and as their size increases they are assigned to latter grids as shown in Fig. \ref{fig:ScaleHeads}. Consequently, each grid learns scale-specific characteristics and features, thus in a way normalizing the learning process.

An important point to consider here is that naively assigning objects to grids based on the actual size normalized with respect to the image can cause imbalancing issues where the first few grids will have a much larger number of boxes assigned to them as shown in Fig. \ref{fig:boxes_assign}, leading to a more complicated learning process. To combat this problem the assignment is instead done using the normalized square root of the object area with respect to the image as outlined in Algorithm \ref{alg:bboxselect}. The assignment algorithm takes the ground truth bounding box location of each object and finds the \textit{x,y} coordinates to map them to. Then it has to select a possible detector from one of the grids. The selection is performed by first evenly splitting the object area space (i.e., IoU values) and assigning a grid to each interval, and then selecting for each object the grid that the square root object IoU falls within. This is facilitated by the $getGridResponsibleArea()$ function. It is essentially a lookup table that returns the area for which each grid is responsible for, which is between $(\dfrac{(g-1)\times(W\times H))}{k}) < boxArea \leq (\dfrac{g\times(W\times H)}{k})$, where $g$ is the index of a grid starting from $1$, and $k$ is the total number of grids. In case of collisions (i.e., bounding boxes mapped on the same grid) the next grid interval is chosen for the assignment. Since this implies that objects occupy the same space or are in perfect alignment with regards to the camera \textit{z} axis the collisions do not happen very often. In such instances they can also be beneficial as they provide some noisy input that can combat over-fitting.

Unlike FPN based approaches \cite{FPN:2016}, detection is not performed on each resolution of the decoder. Instead, each resolution is passed through convolutional layers and resized to the size of the target resolution feature map. A concatenation of these feature maps is then performed. The resulting feature map is forwarded into the detection head to predict outputs (classification and coordinate offsets). Avoiding multi-scale prediction removes ambiguity in ground truth to pyramid level assignment during training.

\subsection{Learning}

\subsubsection{Loss Function}

To learn the pedestrian detection task for the whole model in an end-to-end fashion, a multi-task loss function is employed, combining features from YOLOv2 \cite{YOLOV2}, SSD \cite{liu2016ssd} and LapNet \cite{chabot2019lapnet}. The loss function, shown in Eq. \ref{eq:loss}, is comprised of two components, one for object prediction and one for object localization. 

The localization component (Eq \ref{eq:locloss}) predicts the bounding box offsets. The center box coordinates denoted as $x,y$ are computed similarly to YOLOv2\cite{YOLOV2} where a sigmoid function provides an offset from the starting position of each cell in a grid. The removal of anchor boxes from the overall framework means that the width and height ($w,h$) can be regressed without any priors. Instead these are associated with the predetermined ConvNet input image size. Since they need to be positive they are passed through a sigmoid function and then multiplied with input image width and height ($W,H$) to ensure that the predicted values are within a valid range. Finally, the third component, which is not included in the YOLO \cite{YOLOV2} formulation, is the measure of intersection-over-union (IoU) between the predicted and ground truth boxes. Intuitively the predicted bounding box should perfectly overlap with the ground truth. Hence, to minimize this component we want the IoU to be close to \textbf{1} as shown in Eq. \ref{eq:locloss}. Incorporating the IoU in the loss calculation provides another supervision signal that helps provide better accuracy for object detection \cite{chabot2019lapnet}. Also due to the fact that the IoU is a bounded function it tends to be more stable when predicted offsets must be high. Incorporating all the localization components is critical in obtaining a high accuracy. Also, through an ablation study, in Section \ref{sec:abl}, it is observed that leaving IoU component out, results in reduction in performance. The squared error and the L1-smooth losses\cite{liu2016ssd} (Eq. \ref{eq:l1smooth}) are used to regress the bounding box center and dimensions, and IoU respectively. 

The second component in the loss function relates to object presence (Eq. \ref{eq:oploss}) and class (Eq. \ref{eq:clsloss}). These are similar to YOLO with the difference that the object class is calculated through binary-crossentropy. The object presence score tries to approximate the IoU score for each predicted box. It might seem redundant to keep both terms for single-class detection problems such as pedestrian detection, however, the latter can effectively be used to filter out candidate detections with low object presence score thus reducing the need for post-processing and filtering techniques which can have high complexity.

The responsibility selector $S_{ijk}$ selects the grid $k$ and the detector at position $i,j$ for learning a particular bounding box. It has a value of $1$ for the responsible grid location and $0$ otherwise. Two tuning parameters are present in the loss function. The $\alpha$ value acts as a suppressor for negative boxes and reduces their impact in the loss function computation. This is an important for the training as can hinder the learning process. A value of $0.1$ was empirically found to perform well, lower values also produced good results with an increase in false positives, however. The $\xi$ value controls which of the two loss components, localization, or detection, receives more emphasis. Again, empirically a value of $5$ was found to performing well since the localization loss also indirectly impacts the objectness score. Hence, a higher cost emphasizes that aspect as well. Overall, the multi-component loss has a positive impact on the final accuracy as illustrated by the results.

\begin{align}
\label{eq:loss} & \nonumber Loss = \sum_i\sum_j\sum_k [\xi\times S_{ijk}\times LOC_{loss}(i,j,k) + \\ 
& OP_{loss}(i,j,k)+C_{loss}(i,j,k)]\\\nonumber
\end{align}

\begin{align}
\label{eq:locloss}&\nonumber LOC_{loss}(i,j,k) = smooth_{L1}(x_{ijk}-\hat{x_{ijk}}) + \\
&\nonumber smooth_{L1}(y_{ijk}-\hat{y_{ijk}}) + (w_{ijk}-\hat{w_{ijk}})^2 + \\
& (h_{ijk}-\hat{h_{ijk}})^2 + smooth_{L1}(IOU_{ijk} - \mathbb{1} )\\\nonumber
\end{align}

\begin{align}
\label{eq:oploss}&\nonumber OP_{loss}(i,j,k) = \alpha\times(1-S_{ijk})\times|O_{ijk} - \mathbb{0}| + \\
& S_{ijk}\times|O_{ijk} - IOU_{ijk}|\\\nonumber
\end{align}

\begin{align}
\label{eq:clsloss}&\nonumber C_{loss}(i,j,k) = -[S_{ijk}\times\log(C_{ijk})\\
&+(1-S_{ijk})\times\log(1-C_{ijk})] \\\nonumber\end{align}

\begin{align}
& \nonumber \text{where } S_{ijk}=\begin{cases}
      1, & \text{if grid cell responsible for object}\\
      0, & \text{otherwise}
    \end{cases}\\
&\nonumber \text{and  }\alpha = 0.1, \xi = 5\\
\label{eq:l1smooth} &smooth_{L1}(x) = \begin{cases}
      0.5x^2, & \text{if}\ |x|<1 \\
      |x|-0.5, & \text{otherwise}
    \end{cases}
\end{align}

\subsubsection{Training Process}
Images for training are collected from the PETS2009\cite{PETS2009} dataset, which is aimed for the development of surveillance systems for the detection and tracking tasks within a real-world environment in a variety of scenarios, and different conditions with regards to illumination, viewpoint, occlusion, and crowd density. For each image sequence with ground truth data the full image frames are extracted and smaller patches of size $320\times320$ are cropped which contain the objects and ground truth bounding boxes. Training and test datasets are constructed in this way, both comprised of $1000$ images. 

All the networks are developed and tested through the same framework to have the same conditions and a fair comparison during the inference phase. The Keras deep learning framework \cite{keras}\footnote{Keras version $2.2.0$ with Tensorflow version $1.8.0$} is used which has available all the pretrained models used for transfer learning, with Tensorflow \cite{tensorflow} running as the backend. The same image size is used for all trained networks which was $320\times320$. The weights of the network are normally distributed $ \mathcal{N}(\mu,\,\sigma^{2})$, with $\mu=0$ and $\sigma=0.2$. An $L2$ regularizer is also applied at the convolutional layers with weight decay of $5\times10^{-4}$.

During training heavy image augmentation is applied such as rotations, scaling, noise, blurring, pairwise sampling (mixup), and hue and brightness transformations. The proposed \textit{YOLOpeds} network as well as other methods used for comparison are trained on the same dataset with the same infrastructure. All the networks where trained using a GeForce Titan Xp, on a PC with an Intel $i7-7700K$ processor, and $32$GB of RAM. The SGD optimizer \cite{wilson2017marginal} is used with a learning rate of $0.0001$, with weight decay of $5\times10^{-4}$, and momentum of $0.9$. Each network is trained for $500$ epochs with a batch size of $16$. Also images values are scaled between $[0,\dots,1]$ by dividing the pixels by $255$ for images trained from scratch, while the default preprocessing was applied for transfer learning networks.

\subsection{Inference}
Having the learned parameters of the network and given an input image, an object detector returns a list of bounding boxes and associated confidence scores for each class. The coordinate values are initially between $[0,\dots,1]$. The width and height $w,h$ are then multiplied with the respective factors to obtain the box width and height, and the grid offsets are added to the $x,y$ coordinates to obtain the box center location. A candidate detection is valid if both the object presence and classification scores are over $0.5$. To remove duplicate detections, non-maximum-suppression is applied.

\begin{figure}[t]
\centering
\includegraphics[width=0.7\linewidth]{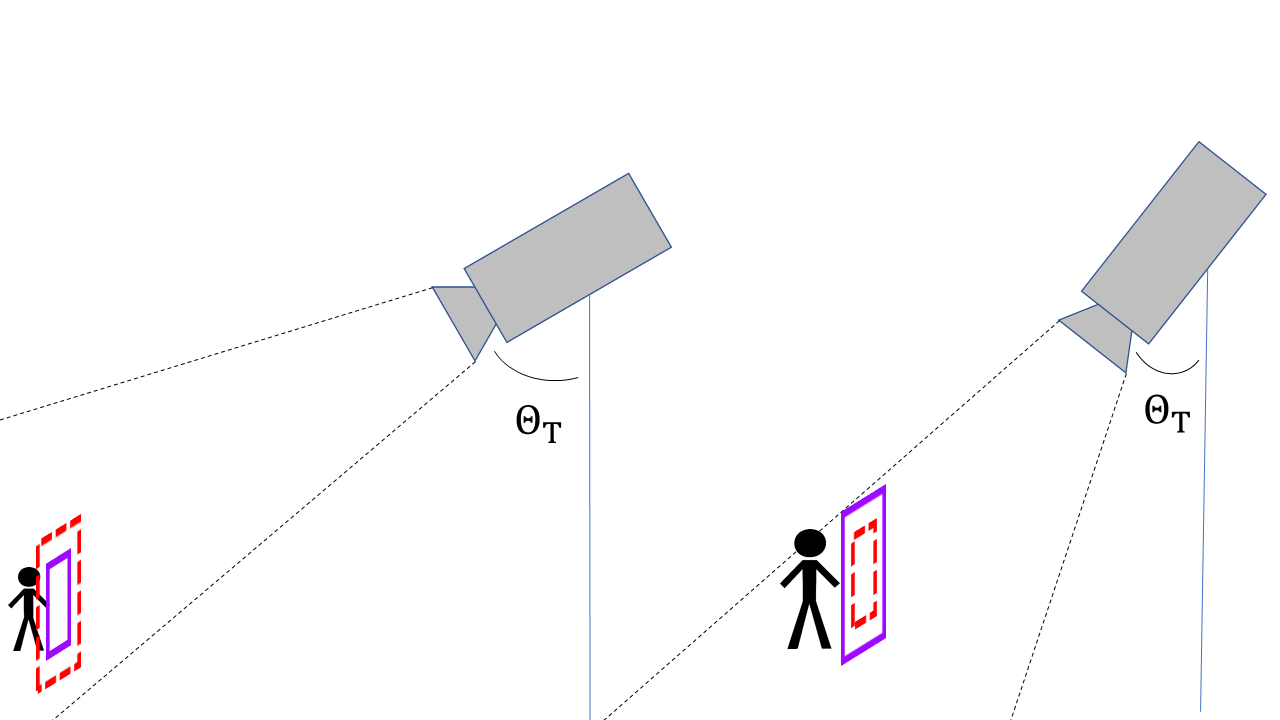}
\caption{Illustration of valid predicted boxes depending on camera positioning e.g., camera tilt angle $\Theta_T$. Blue boxes are valid and dashed red boxes are invalid. (Best viewed in color)}
\label{fig:ptcams}
\end{figure}

The fact that each grid specializes on particular object sizes makes it possible to remove it an informed way by exploiting additional camera parameters and information. For example, in the case of pan-tilt cameras as shown in Fig. \ref{fig:ptcams}, different tilt angles can invalidate certain box scales. Hence, a mapping function $F(G_{ijk},H,\Theta_{T},\Theta_{P})$ that receives the grid predictions $G_{ijk}$ and different parameters such as camera height $H$, and orientation $(\Theta_{T},\Theta_{P})$ that impact the possible distribution of expected boxes can therefore be learned in order to assign scale-specific weights for object presence or even discard entire grids all together, effectively enhancing the proposed approach for active camera applications. 

\section{Experiments}\label{experiments}
In this section, the effectiveness of \textit{YOLOpeds} single-shot framework is demonstrated through various experiments. This section presents experimental results for the application of pedestrian detection and implementation details on an Nvidia Jetson TX2 board \cite{NVIDIA:Jetson:TX2}. The TX2 board includes a Pascal GPU with 256 CUDA cores and 6-core CPU \footnote{Dualcore Nvidia Denver2 and Quad-core ARM Cortex-A57} and supports five alternative modes for the trade-off between speed and energy consumption. Comparisons are made with different backbone networks available in the Keras framework, which are available with weights from ImageNet pretraining and then fine-tuned on the derived PETS2009 dataset with the addition of the \textit{YOLOpeds} detection head. Similarly, the proposed approach is also compared against methods such as YOLOv3 and tinyYOLOv3 \cite{YOLOV3} and SSD\cite{liu2016ssd} which are available with weights from training on the MSCOCO dataset, and which are also fine-tuned on the derived PETS2009 dataset. For all cases, the performance is evaluated on the Nvidia Jetson TX2 Board. All experiments on the Jetson TX2 were carried out using the normal mode of the board without any optimizations or high-power modes. This is more indicative to real applications that need to provide good energy efficiency but operate over longer periods of time, whereas prolonged operation under the high-speed modes can cause overheating and potential damage.

\begin{figure*}[t]
\centering
\includegraphics[width=1\linewidth]{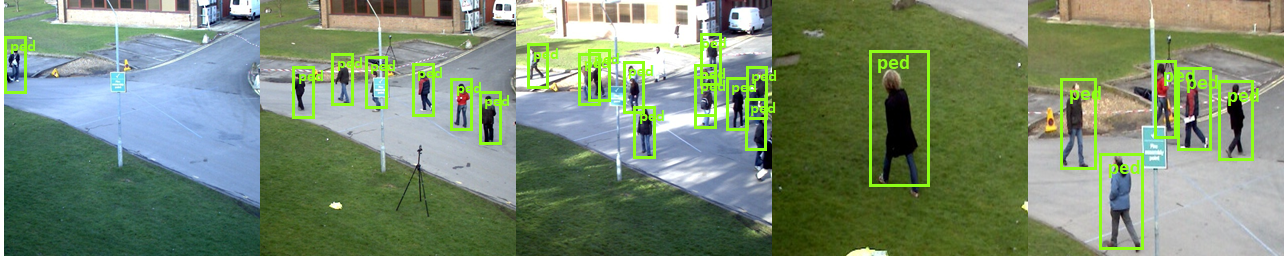}
\caption{Detection results on the PETS2009 image sequences. The proposed approach is able to detect pedestrians at different scales, backgrounds, and densities.}
\label{fig:detres}
\end{figure*}

\begin{figure*}[t]
\centering
\includegraphics[width=1\linewidth]{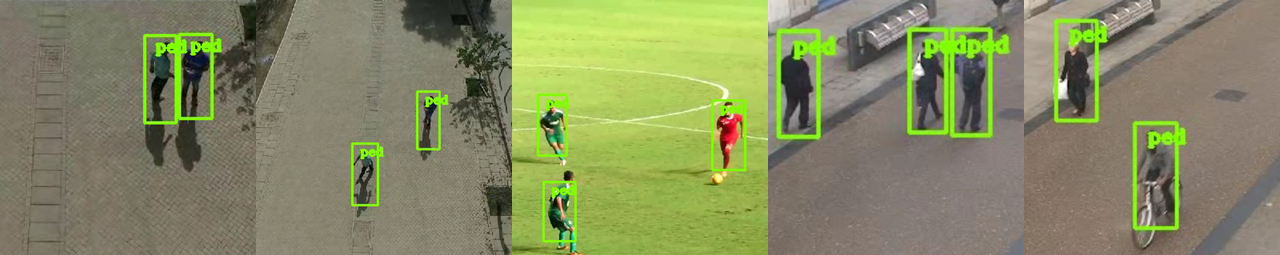}
\caption{Detection results on out of distribution image sequences. Specifically, some images from university campus, football match, and city streets are shown where the proposed network is capable of detecting the pedestrians. (Best viewed in color)}
\label{fig:detres2}
\end{figure*}

\subsection{Metrics and Evaluation}\label{sec:metr}
The evaluation for all networks is performed on the image test set derived from PETS2009. To assess the performance of the detection capabilities common evaluation metrics are used such as intersection-over-union (IoU) and accuracy between correctly predicted labels and ground-truth labels. The different approaches are analyzed and evaluated on the same test dataset using the following metrics:

\textbf{• Accuracy (ACC):}
This metric is defined as the proportion of true positives that are correctly identified by the detector and it is widely used as an accuracy metric, that returns the percentage of the correctly classified objects. Is calculated by taking into account the True Positives ($T^{pos}$) and False Negatives ($F^{neg}$) of the detected objects as given by (\ref{accuracy}).

\begin{equation}
    \label{accuracy}
    ACC = \frac{T^{pos}}{T^{pos}+F^{neg}}
\end{equation}

\textbf{• Intersection over Union (IoU):}
The Intersection over Union (IoU) metric determines the overlap between the ground truth boxes and the positive predictions. Is calculated by considering the intersection between 2 boxes against their union as given by (\ref{iou}). It is averaged over all positive boxes $N_{pos}$

\begin{equation}
    \label{iou}
    IoU = \dfrac{1}{N_{pos}}\times\sum_{i=1}^{N_{pos}}\dfrac{pred\_box_i \cap true\_box_i}{pred\_box_i \cup true\_box_i}
\end{equation}

\textbf{• Frames-Per-Second (FPS):}\label{sec:fps}
The rate at which a detector is capable of processing incoming camera frames, where $t_i$ is the processing time of a single image. 

\begin{align}
	FPS_{model} = \dfrac{1}{\dfrac{1}{N_{test\_samples}}\times\sum_{i=1}^{N_{test\_samples}}(t_i)},  \label{eq:fps}
\end{align}

\textbf{• FPPI (False Positives Per Image):}
FPPI refers to the number of false positive per image given as the fraction of total false positives to the total number of test images. 

\textbf{• Miss Ratec(MR):} The proportion of missed detections (($F^{neg}$)) over all the samples that belong to the target class.
\begin{align}
	MR_{model} = \dfrac{F^{neg}}{F^{neg}+T^{pos}},  \label{eq:mr}
\end{align}

\subsection{Evaluating \textit{YOLOpeds} features}\label{sec:abl}
The main objective of \textit{YOLOpeds} is to provide an accurate detector that is both efficient and fast, thus different components are introduced to improve learning with a smaller network architecture. Next, an evaluation is performed  on the impact by each addition. The most important design decisions that led to \textit{YOLOpeds}, besides the backbone, are shown in \ref{tab:ablation}. The first, concerns the addition of the IoU component in the original YOLO loss formulation. As seen in Table \ref{tab:ablation} this results in a significant improvement over just regressing the bounding box values. It is important to note that removing the direct regression leads to a slight decrease, hence, it is important for the loss to consider all the components. Adding, the short connections also helps in improving the accuracy. The longer connections improve performance even more as they allow the network to look at a diverse set of features to make the predictions. Finally, we attempt to show the impact of the gird size with regards to accuracy. YOLOpeds uses $5$ grids of $10\times10$ which essentially predict a total of $500$ objects and bounding boxes. We evaluate the performance using $5\times5$ grids which essentially predict a total of $125$ objects and bounding boxes. The performance drops by $6\%$. This demonstrates the impact of having enough grid cells so that nearby targets can be better accounted for.

\begin{table}[t]
  \begin{center}
    \caption{Impact of main additions}
    \begin{tabular}{lcccc}
       &  &  &  & \\
      \hline
      IoU loss component &  & \checkmark & \checkmark & \checkmark \\
      \hline
      Short Skip Connection &  &  & \checkmark & \checkmark \\
      \hline
      Long Skip Connections &  &  &  & \checkmark \\
      \hline
       \textbf{Accuracy} & 71.1 & 81.1 & 82.9 & 85.7
    \end{tabular}
    \label{tab:ablation}
  \end{center}
\end{table}

\subsection{Comparison of Networks} \label{sec:comp}
The overall performance of \textit{YOLOpeds} and the proposed single-shot pedestrian detector is presented in Table \ref{tb:compArch}. It is compared against different backbones, all initialized with ImageNet pretrained weights, using the same \textit{YOLOpeds} framework. As seen from the table, the model using the proposed \textit{YOLOpeds} backbone achieves $0.63$ IoU, while having an accuracy of $85.7\%$ higher by $1.5\%$ than the second best backbone which is VGG16. Overall, with regards to false positives and miss rate the \textit{YOLOPeds} backbone performs similarly to other backbones. The ResNet50 model shows the best performance with respect to false positive per image. To investigate if better efficiency can be reached we train a smaller variant ResNet25 which resulted in increased frame-rates with deterioration of FPPI and MR however. The MobileNet archtiectures achieve higher parameter efficiency however, they do so with a cost of degraded accuracy as they do not employ multi-resolution feature fusion. It is worth noting that the same \textit{YOLOpeds} backbone provides good accuracy performance as it utilizes dense connections that result in more diverse feature sets and at the same time performs predictions using information from different resolution feature maps. This demonstrates the effectiveness of multi-scale fusion connections. 

\begin{table}[t]
    \caption{Comparison of Networks in terms of accuracy metrics, performancce on Jetson TX2, and parameter efficiency. }\label{tb:compArch}
\begin{center}
\scalebox{0.9}{
    \begin{tabular}{l|c|c|c|c|c|c|c|c|c} 
      \textbf{Model} & \textbf{Parameters} & \textbf{ACC} & \textbf{IoU} & \textbf{FPPI} & \textbf{MR} & \textbf{APP} & \textbf{FPS} & \textbf{Memory}\\
        &  & (\%)  & (\%) & ($FP/N_{img}$) & & (ACC / MP) & ($1/s$)  & (MB) \\
      \hline
      \hline
      YOLOpeds & $203,899$ & $85.7$ & $63$ & 0.23 & 0.12 & $4.21$ & $33.3$  &  $0.815$ \\
      \hline
      VGG16 & $15,903,040$ & $84.2$ & $60$ & 0.25 & 0.11 & $0.05$ & $5.5$  & $63.6$ \\
      \hline
      ResNet50 & $28,315,008$ & $83.8$ & $60$ & 0.2 & 0.13 & $0.03$ & $7.6$ & $113.26$ \\
      \hline
      ResNet25 & $5,348,224$ & $80.7$ & $60$ & 0.22 & 0.17 & $0.15$ & $15.8$ & $21.3$ \\
      \hline
      MobileNet & $5,596,864$ & $82.8$ & $61.7$ & 0.28 & 0.15 & $0.15$ & $16.6$ & $22.3$ \\
      \hline
      MobileNetV2  & $2,296,384$ & $84.1$ & $62$ & 0.27 & 0.13 & $0.37$ & $20$ & $21.3$ \\
      \hline
      SSD300-MobilenetV2  & $6,368,313$ & $84.4$ & $0.6$ & 0.18 & 0.15 &  $0.13$  & $14.2$  & $25.4$ \\
      \hline
      YOLOv3$^1$  & $61,949,149$ & $87.7$ & $0.65$ & 0.16 & 0.12 & $0.01$  & $4.1$ &  $247.7$ \\
      \hline
      tinyYOLOv3$^1$  & $8,676,244$ & $85.5$ & $0.56$ & 0.39 & 0.14 & $0.1$  & $12$  & $34.7$ 
    \end{tabular}}
    \end{center}
    \text{$^1$Image size of YOLOv3/tinyYOLOv3 is $416$}
\end{table}

An additional benefit of the proposed backbone is that it uses much less parameters than other backbones. An analysis on the parameter utilization is further shown in Table \ref{tb:compArch}. The accuracy contribution per parameter (APP) is used to qualitatively analyze the relationship between detection accuracy and the number of parameters and its unit is percent per million parameters. The \textit{YOLOpeds} backbone boasts an impressive APP ratio of $4.2$ compared to other networks which are less than $1$.

Comparisons are also made with existing generic single-shot detection approaches YOLOv3/tinyYOLOv3\cite{YOLOV3} and SSD\cite{liu2016ssd}. These networks provide weights trained on much larger datasets such as VOC and MSCOCO, and they are further fine-tuned on the same training set\footnote{Default configurations and settings are used for these networks}. The MobileNetV2 variant of the SSD approach is used as it is faster and smaller than its VGG16 variant and thus more suitable for comparisons. We also compare against the YOLOv3 network and its smaller tinyYOLOv3 variant. Overall, they exhibit similar performance as with the other backbones and \textit{YOLOpeds}. YOLOv3 manages to produce the best performance also in terms of false positives per image because it is comprised of a significantly more complex backbone with over 61 million parameters, and also features multi-scale prediction. Both SSD MobileNetV2 and tinyYOLOv3 provide improvements in terms of FPS over YOLOv3 but are still not as efficient as \textit{YOLOpeds}. From the above experimental analysis, we can see that the proposed network structure can achieve good progress in terms of accuracy, parameter utilization and memory usage compared to other detection backbones and existing frameworks.

It is important to note that to scale the approach to more complex data sets such as VOC or COCO it might be necessary to increase the backbone complexity and further tune some of the hyper-parameters. However, it is not the objective of this research to present a generic object detector. Rather, through the proposed evaluation the results are encouraging for developing more efficient application-specific pedestrian detection in smart camera systems.

Some detection results are shown in Fig. \ref{fig:detres} from the PETS2009 test set and Fig. \ref{fig:detres2} from other images. In all cases, good detection performance is observed. Even in conditions where there are shadows and objects such as bicycles the proposed approach manages to detect the pedestrians correctly. There are cases however, where the localization could be improved such as in the football match in Fig. \ref{fig:detres2} case where the box does not cover the entirety of the target.

\begin{figure}[t]
\centering
\includegraphics[width=0.7\linewidth]{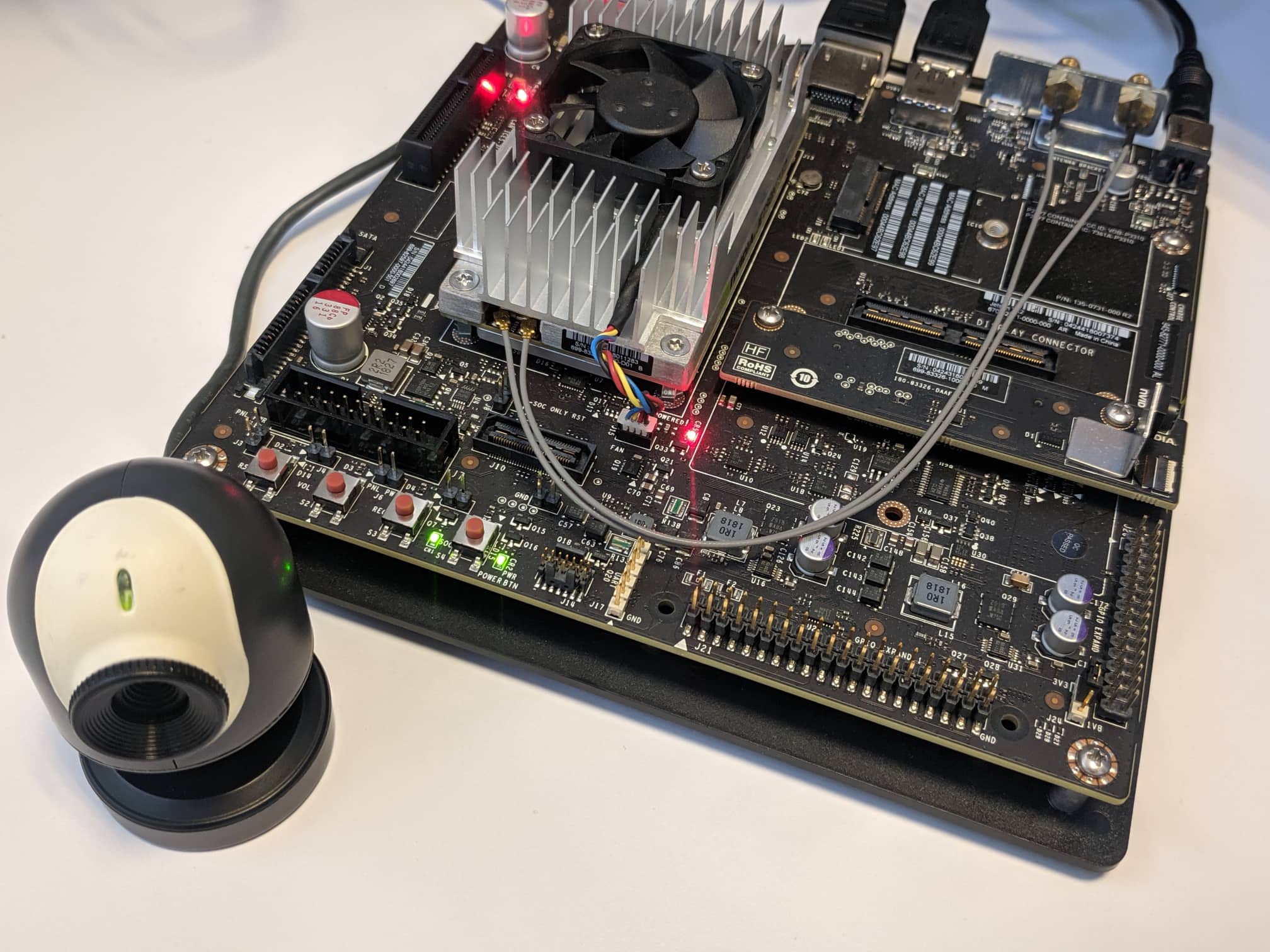}
\caption{Experimental smart camera setup with the NVIDIA Jetson TX2 and a webcam to capture and process live video.}
\label{fig:expsetup}
\end{figure}

\subsection{Performance on Jetson TX2}
Table \ref{tb:compArch} further compares the performance of different networks for system implementations on the Jetson TX2 platform (Fig. \ref{fig:expsetup}). The system using the proposed backbone outperforms the others in terms of frames-per-second. Specifically, it is $1.6\times$ faster than the larger MobileNetV2 and has double the frame-rate from all other networks. Such characteristics make this backbone suitable even for non-GPU platforms. This is also reflected in the memory requirements where it is less than $1$ MB making it suitable for resource constraint devices. One important implementation detail with regards to the inference is that the batch size is $1$. This is because the application scenario is the processing of live camera feed where each frame is consumed prior to moving to the next. In this use-case a batching scheme is not suitable. As can be seen from Table \ref{tb:compArch}, compared to SSD and YOLOv3 the proposed \textit{YOLOpeds} network provides a better inference speed and is much more efficient for memory and computational requirements.

\subsection{Discussion}\label{sec:disc}
Smaller convolutional neural networks have many desirable properties such as reduced computational requirements which makes them faster for both inference and training, they require smaller memory footprint for both storage and working memory, all of which lead to an overall more efficient energy consumption. As such, the proposed approach utilizing a lightweight feature extractor is geared towards that direction and to contribute towards the realization of always-on pedestrian detection in smart camera applications.

Examining the network properties is crucial for understanding why it performs so well and achieves such a high performance in terms of speed and accuracy. Initially, the dense connectivity between the different layers enables it to learn reach representations that are not trivial under the standard sequential paradigm of more traditional ConvNets used for classification such as VGG. Similarly, the combination of multi-scale information from three different network positions makes for a richer set of features on which to regress the bounding boxes. Finally, the specialization of each detection-head grid to a particular set of sizes helps to further improve performance and permits for learning scale-aware properties.

Overall, the proposed feature extraction network in conjunction with the overall framework modifications has achieved a good balance between accuracy, speed, and model size. It boasts an improved accuracy of over $1.3\%$, a speedup of at least $1.6\%$ in terms of framerate compared to the faster available networks, as well as a very high efficiency in terms of accuracy per parameter. Consequently, it can be used in the low-power spectrum for non-GPU devices with limited resources. 

Of course, it is important to expand these results on more diverse datasets but the main motivation behind this work was to demonstrate the benefits of specialized application specific networks. Narrow-domain networks in conjunction with specialized hardware can facilitate the next generation of smart cameras and relevant applications including video surveillance, autonomous driving, robots, and drones. In addition, promising research directions may include the integration of visual attention mechanisms for reduction of input data and the design of lightweight networks geared towards high resolution images.

\section{Conclusion}\label{sec:conc}
In this work, a single-shot pedestrian detection approach referred to as \textit{YOLOpeds}, is presented. Overall, the focus is on providing simplicity in the overall training framework and efficiency with regards to the feature extractor in order to tackle the problems caused by fine-tuning from pretrained networks and generic object detection frameworks. By taking the pretraining-free advantage and not being constraint by an existing backbone pretrained on ImageNet, it is possible to explore a custom architecture for efficient pedestrian detection. Specifically, a densely connected feature extractor with multi-scale feature fusion is proposed. \textit{YOLOpeds} further improves the detection accuracy with respect to existing backbones and provides competitive accuracy with respect to models trained on much larger datasets. Beyond detection accuracy, \textit{YOLOpeds} provides many merits with respect to efficiency by having a reduced number of parameters while offering much improved detection speeds suitable for embedded smart cameras. In addition, exploiting the characteristics of single object detection applications a novel detection head was presented that does not require using anchor-boxes and provides a way to incorporate additional camera information. Overall, the proposed approach can significantly reduce the inference time without diminishing the accuracy, extending the applicability on real-time smart camera applications.

\section*{Acknowledgment}
This work is supported by the European Union\text{'}s Horizon 2020 research and innovation programme under grant agreement No 739551 (KIOS CoE) and from the Republic of Cyprus through the Directorate General for European Programmes, Coordination and Development.
Christos Kyrkou gratefully acknowledges the support of NVIDIA Corporation with the donation of the Titan Xp GPU used for this research.

\bibliographystyle{plain}
\bibliography{references}

\end{document}